\documentclass{article}
\usepackage{spconf,amsmath,graphicx,hyperref}
\usepackage{etoolbox} 

\makeatletter
\patchcmd{\@maketitle}{\large}{\Large}{}{\errmessage{Patching failed}}
\makeatother

\title{UPRPRC: Unified Pipeline for Reproducing Parallel Resources - Corpus from the United Nations}
\twoauthors
  {\parbox[t]{0.9\columnwidth}{\centering 
      Qiuyang Lu$^*$\thanks{$^*$These authors contributed equally to this work.}, Zhengkai Tang, Qiang Liu$^{\dagger}$, \\ 
      Hexuan Cheng, Hui Liu
     }
  }
  {MNBVC Team \\ Global}
  {\parbox[t]{0.9\columnwidth}{\centering
      Fangjian Shen$^*$, Wushao Wen$^{\dagger}$\thanks{$^{\dagger}$Corresponding authors: Qiang Liu \texttt{Q.L.Liu@hotmail.com}, Wushao Wen \texttt{wenwsh@mail.sysu.edu.cn}.}
     }
  }
  {Sun Yat-sen University \\ Guangzhou, China}

\begin{document}

\makeatletter
\let\originaltitle\title 
\renewcommand{\title}[1]{
  \originaltitle{#1 \par \vskip 1em \normalsize \rmfamily \mdseries%
    (DEBUG: Title font size is \texttt{\f@size pt})%
  }
}
\makeatother

\maketitle
\ninept

\begin{abstract}
The quality and accessibility of multilingual datasets are crucial for advancing machine translation. However, previous corpora built from United Nations documents have suffered from issues such as opaque process, difficulty of reproduction, and limited scale. To address these challenges, we introduce a complete end-to-end solution, from data acquisition via web scraping to text alignment. The entire process is fully reproducible, with a minimalist single-machine example and optional distributed computing steps for scalability. At its core, we propose a new Graph-Aided Paragraph Alignment (GAPA) algorithm for efficient and flexible paragraph-level alignment. The resulting corpus contains over 713 million English tokens, more than doubling the scale of prior work. To the best of our knowledge, this represents the largest publicly available parallel corpus composed entirely of human-translated, non-AI-generated content. Our code and corpus are accessible under the MIT License.

\end{abstract}




\section{Introduction}
\label{sec:intro}

Multilingual communication is central to studying and building language technologies for intergovernmental processes. To advance this goal, our open source community, MNBVC \cite{mnbvc}, is creating a large-scale, freely available multilingual corpus. We believe this effort will strengthen multilingual information exchange and research.

The United Nations (UN) is a key source for such data, providing high-quality, human-translated parallel texts in its six official languages. Early releases, like the General Assembly resolutions collection, have already demonstrated the research value of these materials \cite{rafalovitch-dale-2009-united}. We therefore target UN documents and process them into fully aligned parallel resources for broad community use.

However, creating and maintaining large-scale parallel corpora from such dynamic sources presents persistent challenges. Notable prior efforts, such as \emph{MultiUN}, crawled the UN's Official Document System to assemble hundreds of millions of words per language \cite{bechara2024creating, shen2025unaligned, gale1994program, imani2021graph, imani2022graph, gupta2012generic}. This provided sentence-aligned resources at scale, later extended to multilingual alignments \cite{eisele-chen-2010-multiun,chen-eisele-2012-multiun}. Yet, these collections often faced several limitations. They were frequently constrained in scope and scale, susceptible to scraping fragility, and difficult to fully reproduce or extend. These issues underscore the need for larger and more robustly generated corpora.

In parallel, the UN's own translation services have explored and adopted data-driven MT. An early English–Spanish prototype trained on UN material showed promising results. Soon after, the \emph{TAPTA4UN} system was scaled to ten language pairs and entered production use \cite{pouliquen-etal-2012-eamt,pouliquen-etal-2013-tapta4un}. This institutional trajectory highlights the need for modern, reliable, and maintainable UN-based resources that support both research and operations.

To systematically address these challenges, this paper introduces the Unified Pipeline for Reproducing Parallel Resources Corpus from the United Nations (UPRPRC), a unified framework that delivers a state-of-the-art parallel corpus through a fully transparent and extensible workflow. This end-to-end pipeline covers the entire lifecycle, from robust data crawling and document conversion to high-performance text alignment. We introduce a flexible M–N alignment approach, mapping $m$ source units to $n$ target units. Unlike prior UN corpora limited to 4–4 alignments at the sentence level, it operates at the paragraph level, allowing for arbitrary many-to-many merges and splits. A core component is newly designed Graph-based Paragraph Alignment (GAPA) algorithm. GAPA offers superior efficiency and accuracy, enabling us to process larger and more complex documents. We release the entire pipeline as open source, including a minimal standalone example and an optional distributed mode.

Building on the foundational United Nations Parallel Corpus v1.0 \cite{ziemski2016united}, we deliver substantial advancements (Table \ref{tab:tab_compare_v1}). We nearly double the number of fully aligned documents (from 86k to 162k) and more than double the English token count (from 335M to 713M), with a temporal shift toward 2000–2023. Critically, our pipeline is fully open-source. This improves upon the prior version’s partially released code and ensures community can reproduce, verify, and extend the entire process. These gains are enabled by a more efficient alignment algorithm, enhanced table processing, and flexible, paragraph-level M–N alignment.

In summary, our main contributions are:

\begin{table}[t]
\caption{Our work compared to UN Parallel Corpus v1.0}
\centering
\setlength{\tabcolsep}{1pt}  
\begin{tabular}{|r|r|r|}
\hline
\textbf{Feature} & \textbf{Prior work} & \textbf{Our work} \\
\hline
Granularity & Sentence & Paragraph \\
\hline
Open source & Alignment code & Full pipeline \\
\hline
Parse table structure & No & Yes \\
\hline
Time coverage & 1990--2014 & 2000--2023 \\
\hline
Documents & 86,307 & 162,336 \\
\hline
English tokens & 334,953,817 & 713,439,637 \\
\hline
M--N alignment & Up to 4--4 & Arbitrary \\
\hline
Time complexity & \(O(S^2 \cdot L)\) & \(O((R+N)\log N)\) \\
\hline
Space complexity & \(O(S^2 + L)\) & \(O(R + N)\) \\
\hline
Largest file size & 4.6MB (XML) & 19.5MB (text) \\
\hline
\end{tabular}


\par\smallskip
  Complexity symbols: \(S\)=sentences/doc, \(L\)=tokens/sent; \(N\)=total tokens/doc, \(R\)=matching token pairs.
  
\label{tab:tab_compare_v1}
\end{table}





\textbf{A new large-scale UN Parallel Corpus (2000–2023)}, which we believe is the largest publicly available parallel corpus verified to be free of AI-generated content, making it a uniquely valuable resource for training and evaluating MT systems in the current landscape.

\textbf{A complete, end-to-end, and fully open-source pipeline (UPRPRC)}, that is reproducible and extensible, hosted in Github at \url{https://github.com/mnbvc-parallel-corpus-team/UPRPRC/}. Handles complexities such as table structures, addressing long-standing issues in corpus creation.

\textbf{The application of the GAPA algorithm for alignment}, demonstrates its superior time and space complexity allowing the processing of documents at a scale previously infeasible.




\section{File Organization and Format}
\label{sec:format}

The corpus is organized into three levels of granularity, each represented in specific formats, and is detailed below:

\begin{table}[t]
\caption{Statistics for File-level Fully Aligned Corpus}
\centering
\setlength{\tabcolsep}{1pt}  
\begin{tabular}{|r|r|r|}
\hline
\textbf{Lang} & \textbf{Available Files} & \textbf{Number of Tokens} \\
\hline
ar & 160{,}885 & 772{,}312{,}812 \\
\hline
de & 8{,}035 & 18{,}826{,}292 \\
\hline
en & 162{,}550 & 888{,}627{,}227 \\
\hline
es & 160{,}293 & 1{,}030{,}780{,}350 \\
\hline
fr & 162{,}313 & 1{,}018{,}027{,}639 \\
\hline
ru & 161{,}935 & 852{,}676{,}861 \\
\hline
zh & 162{,}583 & 1{,}759{,}231{,}832 \\
\hline
\end{tabular}
\label{tab:table_filewise}
\end{table}

\begin{table}[t]
\caption{Statistics for Bilingual Alignment Corpus}
\centering
\setlength{\tabcolsep}{1pt}  
\begin{tabular}{|r|r|r|}
\hline
\textbf{Lang Pair} & \textbf{Aligned Para.} & \textbf{Number of Tokens} \\
\hline
ar -- en  & 15{,}217{,}906 & 596{,}640{,}108 -- 665{,}376{,}152 \\
\hline
de -- en  & 377{,}681 & 14{,}419{,}368 -- 14{,}683{,}841 \\
\hline
es -- en  & 15{,}967{,}431 & 801{,}346{,}475 -- 678{,}435{,}750 \\
\hline
fr -- en  & 15{,}969{,}753 & 795{,}137{,}952 -- 690{,}028{,}841 \\
\hline
ru -- en  & 15{,}939{,}968 & 634{,}188{,}203 -- 680{,}154{,}733 \\
\hline
zh -- en  & 15{,}331{,}650 & 743{,}513{,}473 -- 668{,}623{,}178 \\
\hline
\end{tabular}
\label{tab:table_dual_lang}
\end{table}

\begin{table}[t]
\caption{Statistics for Paragraph-level Fully Aligned Corpus}
\centering
\setlength{\tabcolsep}{1pt}  
\begin{tabular}{|c|c|c|}
\hline
\textbf{Documents} & \textbf{Paragraphs} & \textbf{English Tokens} \\
\hline
162{,}336 & 16{,}220{,}714 & 713{,}439{,}637 \\
\hline
\end{tabular}
\label{tab:table_paralevel}
\end{table}

{\bf File-level granularity:} This level contains text data converted from original DOC files that were scraped from the website, preserved in JSONL format (available in Hugging Face as \texttt{bot-yaya/rework\_undl\_text}). Each line in the JSONL file represents a document identified by a unique symbol and specifies the language of the file using a two-letter abbreviation. In instances where certain symbols do not encompass all intended languages or files are missing (e.g. due to a 404 Not Found error during scraping), these gaps are filled with empty strings to maintain structural consistency. The statistics for the file-level fully aligned corpus are shown in Table~\ref{tab:table_filewise}.

{\bf Bilingual paragraph-level granularity:} Using the algorithm to be discussed in §\ref{alggapa}, the text from the conversion of DOC to TXT is divided into paragraphs by double new-line characters. Each non-English paragraph is then machine translated into English, and a graph-based alignment algorithm as introduced in §\ref{alggapa} is applied to match each non-English text segment as closely as possible with its English counterpart. This granularity allows for the use of the data in models that do not require extensive contextual token lengths. Additionally, a `Hit Rate' is provided, representing the percentage of the Longest Common Subsequence (LCS) that aligns between the two language segments relative to the total paragraph length. These bilingual corpora are hosted individually on Hugging Face, following a consistent naming convention. The Chinese-English dataset, for example, is named \texttt{bot-yaya/undl\_zh2en\_aligned}, with other languages (e.g., Arabic, Russian) following the same \texttt{undl\_[lang]2en\_aligned} format. Statistics for these bilingual corpora are presented in Table~\ref{tab:table_dual_lang}.

{\bf All-language paragraph-block granularity:} Building upon the bilingual paragraph alignment, this level aggregates the largest connected blocks of paragraphs in all languages. When paragraphs in a specific language are incomplete, adjacent paragraphs are merged to form larger, but still aligned, multilingual paragraph blocks. The resulting corpus is provided as part of the MNBVC corpus (accessible in Hugging Face at \texttt{liwu/MNBVC}), statistics are presented in Table \ref{tab:table_paralevel}. 

This structured approach ensures that the data is versatile enough to support a range of machine translation systems and linguistic research, fostering deeper understanding and utilization of multilingual UN documents.


\section{Methodology}
\label{sec:pagestyle}

\begin{figure}[t]
  \centering
  \includegraphics[width=0.46\textwidth]{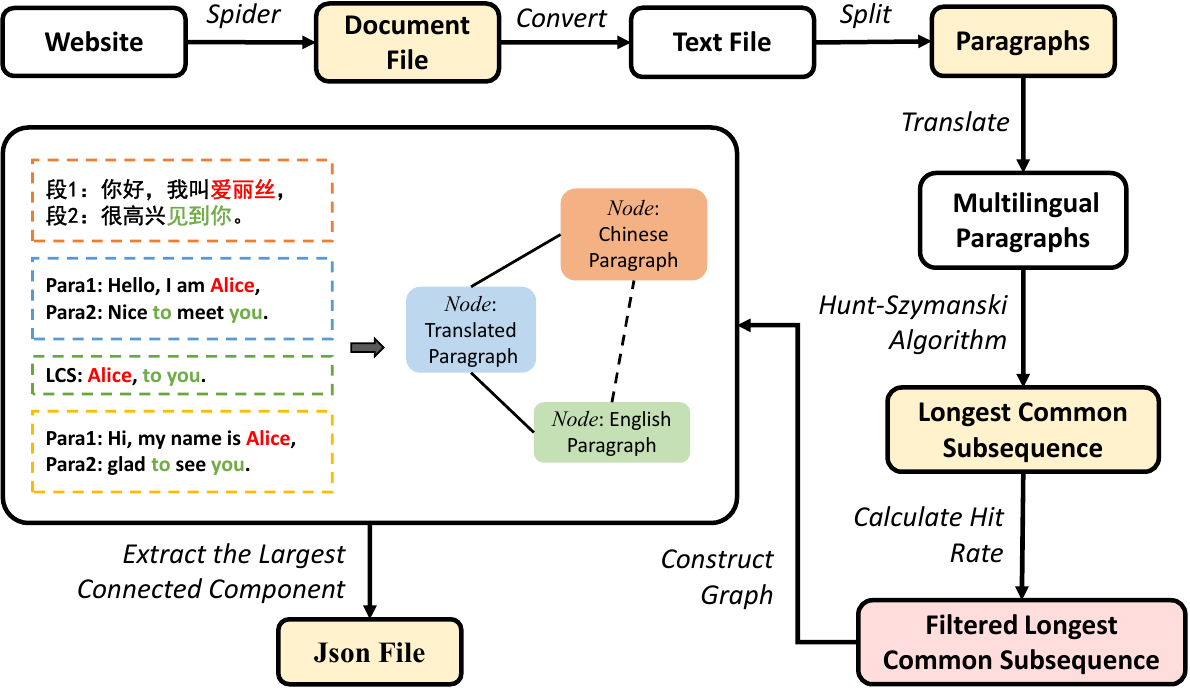}
  \caption{Overview of the complete processing pipeline}
  \label{fig:pipeline-overview}
\end{figure}

This section outlines the sequence of operations within our pipeline, from data acquisition to the preparation of aligned parallel corpora. 
Figure~\ref{fig:pipeline-overview} presents a schematic overview of the entire processing pipeline, illustrating each major stage from initial data acquisition to the production of aligned parallel corpora.

\subsection{Data Collection}
Our initial step in constructing the corpus involved collecting data directly from the United Nations Digital Library \cite{mann2018united}. Due to several updates to the online query service, including an overhaul that deprecated the old APIs, we adapted our approach to utilize the most recent API available at \url{https://search.un.org/search?collection=ods}. The Official Document System (ODS) is pivotal as it offers documents in multiple languages, essential for compiling parallel corpora.

\subsection{Document Download and Verification}
Once we obtained a list of documents, the download from the site was straightforward. However, it is important to note that not all files are in DOC format; some are PDFs. Our attempts to programmatically extract clean paragraph text from PDFs were unsuccessful, leading us to skip these files through file header verification.

\subsection{Document Conversion}

Most of files were in the legacy DOC format, fewer in DOCX and occasional WordPerfect-exported files (.wpf). To facilitate downstream scripting, we standardized all documents to DOCX by batch-converting every non-DOCX file with Microsoft Word 2019 via COM automation. We implemented robust retry and timeout mechanisms to handle anomalies, such as corrupted, encrypted, or unusually large files that could crash Word or trigger excessive memory usage. Files that could not be automatically converted were flagged for manual review, as they may still contain salvageable content. After conversion, the texts were extracted using Pandoc \cite{MacFarlane_Pandoc}, separating the paragraphs with double line breaks.

\subsection{Data Table Processing Workflow}

Before paragraph-level alignment, we normalize and flatten tables to prevent layout artifacts from disrupting boundary detection. After conversion to \texttt{.docx}, we obtain plain text with Pandoc (using \texttt{--wrap=none}) to preserve line structure for downstream parsing. We then strip zero-width and other format-control characters (e.g., U+200E and soft hyphens) to stabilize width calculations. Next, we detect three ASCII-style table patterns commonly produced by office exporters and plain-text writers—(a) multi-line tables with repeated dash rules acting as header/splitter/footer; (b) multi-line tables with only top/bottom delimiters; and (c) grid tables using “+” and “|” borders. For each detected table, column boundaries are parsed from the delimiter lines; cell contents are reconstructed with a Unicode-aware character-width function and replaced by a single inlined line per row (cells space-separated) to yield sentence-like spans. A recursive pass applies these detectors until no tables remain. The flattened paragraphs are written to a dedicated output directory and then fed into the alignment stage in §\ref{alggapa}.

\subsection{Machine Translation and Alignment: Graph-Aided Paragraph Alignment}
\label{alggapa}

The original corpus is aligned at the document level between non-English and English: for each non-English document, there is a corresponding English document and vice versa. We improve the alignment of the corpus from document-level to paragraph-level. However, the number of paragraphs is not equal for each pair of non-English and English documents, and there is no natural correspondence between non-English and English paragraphs. For example, a non-English paragraph $x_i$ may correspond to two English paragraphs $y_j$ and $y_{j+1}$, that is, the $i$-th paragraph of the non-English document contains and only contains the content of the $j$-th and $j+1$-th paragraphs of the English document. Thus, we cannot align two documents na\"ively through ascending order. In this section, we introduce our Graph Aided Paragraph Alignment algorithm (GAPA) to obtain a paragraph-level aligned corpus. 

For each non-English document $x=[x_1, \cdots, x_m]$ and its corresponding English document $y=[y_1, \cdots, y_n]$, where $x_i$ or $y_i$ denotes the $i$-th paragraph of the corresponding document, our goal is to find the mapping list $\mathcal{F}=[F_i, \cdots, F_{\min(m,n)}]$. Each element in $\mathcal{F}$ is $(x_i,[y_j,\cdots,y_{j+k}])$ or $([x_i,\cdots, x_{i+k}],y_j)$. The former means that the $i$-th non-English paragraph corresponds to the $j$-th through $j+k$-th English paragraphs and $[y_j,\cdots,y_{j+k}]$ can be merged into a single paragraph for downstream applications. The latter form $([x_i,\cdots, x_{i+k}],y_j)$ can be understood in a similar way. 

GAPA maps two corresponding documents $x$ and $y$ to a bipartite graph $G(x,y)$ where each node in the graph is a paragraph \cite{bondy1976graph}. If two paragraphs $x_i$ and $y_k$ correspond, then there is a link $(i,k)$ connecting node $x_i$ and $y_k$. Moreover, each map $F_i\in\mathcal{F}$ is a subset of the link list of the bipartite graph $G(x,y)$ and defines a connected subgraph. For example, $(x_i,[y_j,\cdots,y_{j+k}])$ is a link set $\{(i,j),\cdots,(i,j+k)\}$.

The alignment of the paragraphs is reduced to find all links in the bipartite graph $G(x,y)$ and to detect all connected subgraphs. To find all links in the graph, we first process the non-English document by machine translating it into English and then find the Longest Common Subsequence \cite{paterson1994longest} $\mathrm{LCS}(x,y)=[z_1,z_2,\cdots,z_o]$, where $z_i$ is the $i$-th common word, between the translated non-English document $x$ and the corresponding English document $y$. For each common word $z_i$, we find the first two words $w_x=z_i$ and $w_y=z_i$ of the documents $x$ and $y$, respectively. Assume $w_x\in x_i$ and $w_y\in y_j$, which means that the first word $w_x=z_i$ in the translated non-English document $x$ is located in the $i$-th paragraph and the first word $w_y=z_i$ is located in the $j$-th paragraph of $y$, and then we can connect the nodes $x_i$ and $y_j$, that is, find a link $(i,j)$ in the bipartite graph $G(x,y)$. Each word $z_i\in \mathrm{LCS}(x,y)$ creates one and only one link based on the first-word-match rule (fwmr). After processing all words $z_i\in\mathrm{LCS}(x,y)$, all links in the bipartite graph $G(x,y)$ are found. Each connected subgraph $G_i(x,y)\subset G(x,y)$ defines a correspondence of paragraphs in $x$ and $y$. We merge multiple paragraphs into one paragraph denoted by nodes in $x$ and in $y$, respectively, obtaining a correspondence pair of the non-English and English paragraphs.

In real scenarios, the performance of GAPA degenerates due to noisy links connecting nodes $x$ and $y$. For example, the frequent common words $a$, $the$, $is$ may form links between unrelated paragraphs across documents (nodes) $x$ and $y$ \cite{imani2021graph, gupta2012generic, paetzold2017massalign}. To mitigate noisy links, for each paragraph (node) $x_i\in x$ or $y_i\in y$, we introduce the LCS hit rate, which is:

\begin{equation}
h(x_i):= \frac{\sum_{w\in \mathrm{LCS}\cap_{\mathrm{fwmr}} x_i}\mathrm{L}(w)}{\sum_{w\in x_i}\mathrm{L}(w)}
\end{equation}

where $\mathrm{L}(w)$ is the number of letters in the word $w$. The LCS hit rate is the ratio between the total number of letters in the LCS words in the paragraphs $x_i$ and the total number of letters in the paragraph $x_i$. If the LCS hit rate of a paragraph (node) $x_i\in x$ or $y_i\in y$ is below a threshold $h_c$, then we remove the links connecting to the node denoting the paragraph and thus the node is discarded. In the final paragraph-level aligned corpus, GAPA may discard partial paragraphs of the original corpus. There is a trade-off between mitigating noise and maintaining completeness of the original corpus, which is controlled by the LCS hit rate threshold $h_c$. Based on our experiments, $h_c=0.3$ is a suitable threshold.

We adopt Argos for machine translation, an open-source offline Neural Machine Translation (NMT) library. We chose Argos because it is easy to self-host (Python library/CLI/GUI), automatically splits paragraphs into sentences before translation, so that users need not manage chunking or model context-length limits, and it runs fully on local hardware, avoiding the per-request quotas and fees typical of cloud MT APIs. We recorded the translated paragraphs for downstream alignment, and provide the computational cost of this step—provide an optional distributed implementation for multi-machine setups \cite{Finlay_Argos_Translate}.

GAPA is based on the Longest Common Subsequence (LCS), adjusted for paragraphs of different lengths across languages. This method involved complex calculations where traditional dynamic programming was inefficient; hence, we employed the Hunt-Szymanski algorithm \cite{hunt2017hunt}, which is better suited for such tasks.


\section{Experiments}
\label{sec:typestyle}

\subsection{LLM-based evaluation and link to the method hyperparameter}
For the bilingual paragraph-level aligned corpus, we evaluate alignment quality with large language models (LLMs) \cite{naveed2025comprehensive, zhao2023survey}. For each non-English$\rightarrow$English pair (es–en, zh–en, fr–en, ru–en, ar–en), we first discard any candidate whose English side has $<!32$ characters \emph{and} $<!5$ words, as such very short segments are dominated by boilerplate or formulaic expressions and thus produce many duplicate sentences in the short-length stratum—scoring identical samples is uninformative. From the remainder, we take the 100 pairs with the longest English segments and the 100 with the shortest, then uniformly sample 1{,}800 additional pairs—yielding exactly 2{,}000 pairs per language and 10{,}000 pairs per threshold. Each pair is labeled by three LLMs (ChatGLM3~\cite{glm2024chatglm}, GPT-4~\cite{achiam2023gpt}, and Qwen2~\cite{qwen2}) with a binary \emph{True/False} decision for “correct alignment”.


\subsection{Document-level aggregation (definition of \emph{Accuracy})}
To aggregate the pair-level judgments into a more robust quality signal, we introduce a document-level accuracy metric. Let $\mathcal{S}$ be the set of the sampled paragraph pairs for a given threshold \(h_c\), and let $\mathcal{D}$ be the set of \emph{unique source documents} from which these pairs were drawn (multiple sampled pairs may come from the same document). For a given model $M$, define a document to be \emph{good} if \emph{all} of its sampled pairs are labeled \emph{True} by $M$; otherwise, it is \emph{bad}. The reported \emph{Accuracy} is the fraction of good documents:
\[
\mathrm{Acc}_M(h_c)
\;=\;
\frac{\big|\{\, d\in\mathcal{D} \;:\; \forall\,(p,q)\in\mathcal{S}_d,\; \hat{y}_M(p,q)=\mathrm{True}\,\}\big|}
     {\big|\,\mathcal{D}\,\big|},
\]
where $\mathcal{S}_d\subseteq \mathcal{S}$ are the sampled pairs originating from document $d$, and $\hat{y}_M(p,q)\in\{\mathrm{True},\mathrm{False}\}$ is the model’s label for pair $(p,q)$. Intuitively, if any sampled pair from a document is judged misaligned, that entire document is counted as \emph{bad} and excluded from the numerator. Table~\ref{tab:alignment-accuracy} reports this strict, document-level metric aggregated over all five language pairs for each \(h_c\) (DROP\_THRESHOLD) in GAPA.

\begin{table}[t]
\caption{Alignment accuracy judged by model (document-level, “any-error-is-bad”}
\begin{center}
\texttt{DROP\_THRESHOLD} $\,\equiv\, h_c$
\begin{tabular}{|c|c|c|}
\hline
\texttt{DROP\_THRESHOLD} &  Model & Accuracy\\
\cline{1-3} 
& ChatGLM3   & 99.034\% \\
\cline{2-3} 
0.0 & GPT-4      & 93.818\%  \\
\cline{2-3} 
& Qwen2      & 98.068\% \\
\cline{1-3} 
& ChatGLM3   & 98.942\% \\
\cline{2-3}
0.1& GPT-4      & 93.945\% \\
\cline{2-3}
& Qwen2      & 97.931\% \\
\cline{1-3} 
& ChatGLM3   & 99.012\% \\
\cline{2-3} 
0.2 & GPT-4      & 94.023\%  \\
\cline{2-3} 
& Qwen2      & 98.023\% \\
\cline{1-3} 
& ChatGLM3   & 99.012\% \\
\cline{2-3} 
0.3 & GPT-4      & 94.104\%  \\
\cline{2-3} 
& Qwen2      & 98.081\% \\
\cline{1-3} 
& ChatGLM3   & 99.014\% \\
\cline{2-3} 
0.4 & GPT-4      & 94.190\%  \\
\cline{2-3} 
& Qwen2      & 97.994\% \\
\hline
\end{tabular}
\label{tab:alignment-accuracy}
\end{center}
\end{table}

\subsection{Human vs LLM Alignment Audit (English-Only)}

To calibrate the LLM-based labels and validate their reliability, we conducted a human audit focused on the operating point selected for our final pipeline. For this audit, we used only the English side of the data, as alignment issues are often language-agnostic. Specifically, from the 2{,}000 en–side paragraph pairs produced by the sampler at \(\texttt{DROP\_THRESHOLD}\equiv h_c=0.3\) (the hyperparameter used in our final pipeline), we uniformly sampled 100 pairs for human annotation. Table~\ref{tab:humanvsai} reports the counts of disagreements between human ground truth and each LLM judge: for each model, the first row gives false positives (model says aligned, human says misaligned) and the second gives false negatives (model says misaligned, human says aligned). No other languages are included in this audit.

\begin{table}[t]
\caption{Human vs AI}
\begin{center}
\begin{tabular}{|c|c|c|c|}
\hline
Model & Ground Truth & Model Eval Result & {Count} \\
\cline{1-4}
ChatGLM3 & Negative & Positive & 2 \\
 \cline{2-4} 
 & Positive & Negative & 0 \\
 \cline{1-4}
Qwen2 & Negative & Positive & 2 \\
 \cline{2-4} 
 & Positive & Negative & 0 \\
 \cline{1-4}
GPT4 & Negative & Positive & 2 \\
 \cline{2-4} 
 & Positive & Negative & 5 \\
\hline
\end{tabular}
\label{tab:humanvsai}
\end{center}
\end{table}




\section{Conclusion}
\label{sec:majhead}

In this work, we present a complete data‐production pipeline built upon prior efforts, with targeted improvements in the processing of tabular data and a broadly applicable fine‐grained alignment algorithm leveraging large‐scale machine translation. Our pipeline scripts are publicly available on GitHub and are continuously updated, enabling newly released, high‐quality translations from the United Nations to be ingested and converted into machine‐readable datasets in near real time. The resulting large volume of parallel data is poised to play a significant role in the ongoing training and refinement of modern large language models.

Moreover, we periodically publish our curated United Nations parallel corpus on Hugging Face, ensuring that downstream users have immediate access to the most up‐to‐date, high‐quality multilingual data. By providing both the processing tools and the continuously refreshed dataset, we aim to support a wide range of research and applications in multilingual natural language processing.

\bibliographystyle{IEEEbib}
\bibliography{strings,refs}

\end{document}